\pdfoutput=1

\documentclass[11pt]{article}

\usepackage{acl}

\usepackage{times}
\usepackage{latexsym}

\usepackage[T1]{fontenc}

\usepackage[utf8]{inputenc}

\usepackage{microtype}

\title{Automatically Evaluating Opinion Prevalence in Opinion Summarization}

\author{Christopher Malon \\
  NEC Laboratories America \\
  4 Independence Way \\
  Princeton, NJ 08540 \\
  USA \\
  \texttt{malon@nec-labs.com} \\}

\newcommand{\abs}[1]{\vert#1\vert}

\begin{document}
\maketitle
\begin{abstract}
When faced with a large number of
product reviews, it is not clear that a human can remember all of them
and weight opinions representatively to write a good reference summary.
We propose an automatic metric to test the prevalence of the opinions
that a summary expresses, based on counting the number of reviews that
are consistent with each statement in the summary, while discrediting
trivial or redundant statements.  To formulate this opinion prevalence metric,
we consider several existing methods to score the factual consistency of
a summary statement with respect to each individual source review.
On a corpus of Amazon product reviews, we gather multiple human judgments
of the opinion consistency, to determine which automatic metric best
expresses consistency in product reviews.  Using the resulting opinion
prevalence metric, we show
that a human authored summary has only slightly better opinion prevalence than
randomly selected extracts from the source reviews, and previous extractive
and abstractive unsupervised opinion summarization methods perform worse
than humans.  We demonstrate room for improvement with a greedy construction
of extractive summaries with twice the opinion prevalence achieved by humans.
Finally, we show that preprocessing source reviews by
simplification can raise the opinion prevalence achieved by existing
abstractive opinion summarization systems to the level of human performance.
\end{abstract}

\section{Introduction}

Opinion summarization has emerged as a commercial application
of multi-document summarization, with the goal of outputting the
most salient opinions expressed in a collection of customer reviews
of a given product or service \citep{sigir}.  Practitioners have recognized the
difficulty of obtaining a large training set of summaries that adequately
represent a large number of opinions of various products, and so
have developed unsupervised systems for opinion summarization
\citep{brazinskas-etal-2020-unsupervised,
iso-etal-2021-convex-aggregation,
isonuma-etal-2021-unsupervised,
angelidis-etal-2021-extractive}.

Although systems may be trained without reference summaries,
they are still central to system evaluation.
However, for large scale data, adequate human references
may not be only {\em expensive} but actually {\em impossible},
because the human cannot remember all the source text at once.
A commonly used test set of Amazon data summarizes only sets of eight
product reviews \citep{brazinskas-etal-2020-unsupervised}.
The newer SPACE dataset collects summaries for sets of
one hundred product reviews, by dividing the work of selecting important
statements among different annotators, and combining the selected statements
in an aggregation step \citep{angelidis-etal-2021-extractive}.
Even if this is adequate for a hundred reviews, it will not scale when
a human has to combine even more statements.
The largest dataset, AmaSum \citep{brazinskas-etal-2021-learning},
uses the opinions of professional reviewers as references,
which utilize many sources of information including the writers'
own product tests, and may not be based on opinion counting.

For single-document summarization, researchers have introduced reference-free
metrics
to quantify factual consistency of a summary with its source document
\citep{ernst-etal-2021-summary, fabbri-etal-2022-qafacteval,
laban-etal-2022-summac},
but factuality still is largely neglected in multi-document and opinion
summarization.  We argue that a statement in an opinion summary should not only
be factual in the sense of being logically implied by one document (review),
but prevalent in the sense of being logically implied by many documents.

The main contribution of this paper is to introduce an automatic,
reference-free metric
for opinion prevalence in product reviews.  Additionally, we present a 
new dataset, ``ReviewNLI,'' of human consistency judgments in product reviews.
We compare automatic metrics for opinion consistency on ReviewNLI, choosing
the best to form the basis of our opinion prevalence metric.  We quantify the
advantage of human summaries above randomly selected review extracts, but show
that it is possible to write summaries with twice the opinion prevalence of a
human reference summary.  Finally, we introduce a preprocessing technique that
improves the opinion prevalence of some abstractive opinion summarization
systems to human levels.

The ReviewNLI data, code for our opinion prevalence metric, and code for our
preprocessing technique are released.\footnote{\tt https://github.com/cdmalon/\\opinion-prevalence}

\section{Related Work}

Some work on opinion summarization assumes that a review is annotated
with ratings for each {\em aspect} within a finite set,
that aspect seed terms are chosen,
or that an aspect-based sentiment analysis model is available, and
coverage of these aspects and sentiments guides the structure of the
output summary \citep{fabbrizio1, angelidis-lapata-2018-summarizing, suhara-etal-2020-opiniondigest}.
However, even if such information is available,
within each aspect and/or sentiment, the
task remains of aggregating the source review texts relevant to that aspect or
sentiment into part of the summary.  We seek to measure the degree
to which a summary reflects the source reviews.  We quantify this in
a general way that does not require specific annotations.

Opinion prevalence is not measured in recent opinion summarization work,
but some notions of consistency are.
\citet{noisypairing}
automatically measured the consistency between source reviews and the generated
summary based on the similarity of the contextual embedding.
This is a weaker relationship than logical entailment, and
score can be earned for alignment to any source opinion, whether it
is frequent or infrequent.
\citet{suhara-etal-2020-opiniondigest} and
\citet{isonuma-etal-2021-unsupervised} manually evaluated
{\em faithfulness} of their summaries against source reviews.  However,
they formulate faithfulness as a ternary classification problem
(fully, partially, or not supported) which considers a summary
fully faithful if even one source review supports it.
Standards for labeling
and agreement with the majority decisions were not reported.
Recently \citet{hercules} evaluated faithfulness automatically with SummaC
\citep{laban-etal-2022-summac},
again on the basis of whether any supporting review existed, without regard
to the frequency of opinions.

Similarly to other summarization tasks, it is common to manually evaluate
fluency, coherence, informativeness, and redundancy of an opinion summary
\citep{isonuma-etal-2021-unsupervised, brazinskas-etal-2020-unsupervised,
amplayo-lapata-2020-unsupervised}.
We do not suggest that opinion prevalence should replace these.
Other desiderata have been suggested as well.
For example, \citet{noisypairing}  writes that systems may ``reflect
the important opinions in reviews, {\em e.g., This is a nice
place to eat.  The staff are nice and friendly.}, but they may not
generate summaries that grab the attention of their readers''
and attempts to match the style of individual reviews in the summary,
though this is evaluated qualitatively.
A reader might find the style of an anecdotal review more pleasant to read,
but particular anecdotes generally occur only once.
There may be tradeoffs between opinion prevalence and desiderata
such as style matching, and we leave that investigation for future work.

As one construction of extractive summaries with high opinion prevalence,
we use a greedy search strategy similar to the one introduced
by \citet{saggion} to extract summaries that optimized ROUGE. 
Later research used an A* search to extract summaries
with even better ROUGE recall \citep{aker-etal-2010-multi}, but for
opinion prevalence there is interaction among added sentences
to measure redundancy, so such a search may be inefficient.

\section{Automatic Metrics for Opinion Consistency}

\subsection{Data}

To select among possible automatic metrics for opinion consistency,
we used the Amazon input source reviews and human summaries
collected in \citet{brazinskas-etal-2020-unsupervised}.
This dataset contains a development set of 28 products and
a test set of 32 products, each with eight customer reviews in English.
Three human-authored reference summaries appear for each product,
for use in evaluation.  These reference summaries have been
used to evaluate several abstractive unsupervised opinion summarization
systems, including \citet{brazinskas-etal-2020-unsupervised},
\citet{iso-etal-2021-convex-aggregation}, and
\citet{isonuma-etal-2021-unsupervised}.

We use the Punkt tokenizer of NLTK 3.6.2\footnote{{\tt www.nltk.org}} to
split the reference summaries into sentences.
Pairing each sentence from the first reference summary for each product
with each of the source reviews, we asked qualified
crowdworkers to judge whether the sentence was mostly implied or not
by the given customer review (a binary decision).
Details about the instructions and
qualification procedure are given in Appendix~\ref{sec:crowdworker}.
We obtained decisions from three crowdworkers for each sentence/review pair,
and took the majority decision as ground truth.
We release these decisions as the dataset, ``ReviewNLI.''

There are several reasons this opinion consistency judgment
may be subjective.  The same basic issue with a product may be described
with slightly different details.  Moreover, a single sentence of a summary
may combine various pieces of information, only some of which appear
in a given source review.  Through the instructions, examples, and qualification
test, we strived to achieve a mutual understanding of how to resolve
these ambiguities.  For instance, we clarified that a source review
that said ``We liked the clean rooms'' would not entail a summary that said
``The hotel room was clean and bright'' because the rooms may have been
clean but very dark.

\begin{table}[htb]
\begin{center}
\begin{tabular}{lccc}
\hline
Dataset & Accuracy & FP & FN \\
\hline
SNLI & .9228 & .0596 & .1115 \\
MultiNLI & .9293 & .0519 & .1048 \\
ReviewNLI & .9220 & .0621 & .1106 \\
\hline
\end{tabular}
\caption{Worker agreement with majority labels: accuracy, false positive rate on negative examples, and false negative rate on positive examples.}
\label{tbl:agreement}
\end{center}
\end{table}

Through these efforts, we achieved a rate of agreement between worker
labels and the majority decision that is on par with popular
mainstream natural language inference (NLI) datasets.
Table~\ref{tbl:agreement} compares overall labeler accuracy with respect to
the majority, and the false positive and false negative rates, across
our dataset (ReviewNLI), and the development sets of
SNLI \citep{bowman-etal-2015-large}
and (matched) MultiNLI \citep{williams-etal-2018-broad}.
For fair comparison, SNLI and MultiNLI
are relaxed to binary problems, classifying entailment versus
non-entailment (neutral or contradiction).  Labelers achieve between 92\%
and 93\% accuracy on all datasets.  On ReviewNLI, this accuracy was fairly
uniform across workers: each worker who contributed more than 100 labels
had at least 90\% accuracy.

As in SNLI and MultiNLI, roughly one third of the labels in ReviewNLI
are entailments (627 of 1920).

\subsection{Experiments}

We evaluate how accurately existing metrics for summary-source consistency
can be thresholded to predict the human judgments of opinion consistency
found in ReviewNLI.  None of the metrics has been trained on data
from the review domain.

{\bf ROUGE} \citep{lin-2004-rouge} is a popular, model-free
metric.
Here, to test summary-source consistency, we take the ROUGE-1 precision
rather than
recall or F1, as we expect words of the summary sentence to be mostly
contained in the source review if entailed, whereas most words in the source
review are not expected in any one summary sentence.  We apply
stemming and use the implementation in Huggingface Datasets 2.3.2.\footnote{
{\tt github.com/huggingface/datasets}}

{\bf SuperPAL} \citep{ernst-etal-2021-summary} is proposed as a
semantic alternative to ROUGE for measuring summary-source alignment,
by using OpenIE information extraction \citep{stanovsky-etal-2018-supervised}
to extract propositions from summaries and sources, and using a
RoBERTa \citep{roberta}
model first fine-tuned on MultiNLI and further fine-tuned
to decide which pairs of propositions align.
We consider taking the maximum or average alignment score over all
proposition pairs generated for a source review / summary sentence pair.

{\bf QAFE} \citep{fabbri-etal-2022-qafacteval} generates question / answer
(QA) pairs based on a summary sentence, and compares the answers
expected from the summary sentence to the answers predicted by a QA model
using the source review as context.  We take the F1 score over the QA pairs
generated as the measure of consistency.

{\bf SummaC} \citep{laban-etal-2022-summac} combines NLI scores obtained
across units of a specific granularity between a source and a summary.
We use the zero-shot system, taking entailment probability without subtracting
contradiction probability.  After validating different models and granularities
on the development set, we found that the MultiNLI (Roberta large) model with
document granularity had the best performance.

As our task is unbalanced binary classification, for reasons similar to those
discussed in \citep{laban-etal-2022-summac}, we choose balanced accuracy
to select the best metric, while also reporting AUC \citep{bradley}.
Balanced accuracy is defined in terms of true positives (TP),
true negatives (TN), false positives (FP), and false negatives (FN) as
$$\frac{TP}{2(TP+FN)} + \frac{TN}{2(TN+FP)} \ldotp$$
For each method, the best threshold is computed on the
development set and the corresponding balanced accuracy on the test set
along with the AUC is reported in Table~\ref{tbl:prediction}.

\begin{table}[htb]
\begin{center}
\begin{tabular}{lcc}
\hline
Method & Balanced & AUC \\
& Accuracy & \\
\hline
ROUGE & .6501 & .7100 \\
SuperPAL (Max) & .7178 & .7635 \\
SuperPAL (Avg) & .7249 & .7851 \\
QAFE & .5956 & .6073 \\
SummaC (Sent) & .7263 & .7793 \\
SummaC (Doc) & {\bf .7973} & {\bf .8626} \\
\hline
\end{tabular}
\caption{Predicting majority human judgment of opinion consistency
on ReviewNLI.}
\label{tbl:prediction}
\end{center}
\end{table}

SummaC performs the best.  Its best balanced accuracy is achieved
at a threshold which judges summaries as consistent if the
model predicts even a weak (.04) probability of entailment,
indicating that the consistent pairs in the ReviewNLI task are related
in a much weaker way than strict logical entailment.
We observed a 7\% difference between balanced accuracy at the document and
sentence granularities.  At sentence granularity, a maximum was taken
among sentence-based NLI judgments, and this did not perform well
when information was split across source sentences.

Although SuperPAL also leverages an NLI classifier, it did not perform
as well as SummaC, suggesting that its extracted propositions were not
useful enough to classify the relation of summary and source review.
This held even though we tried two ways (maximum and average) of aggregating
the proposition pair judgments.

ROUGE performs worse than every method except QAFE.
Even at its best balanced accuracy, QAFE only classifies 21.5\% of
majority-judged consistent opinion pairs as consistent.
We suspect that QAFE is too factoid oriented to be useful
in confirming opinions,
and this is confirmed by qualitative examples of QA pairs such as
``What type of shoes run a bit narrow?'' which focus on details of the
shoes that should be shared by all instances of the same product,
instead of the conclusion that the shoes are narrow.

\section{Scoring Opinion Prevalence}

Given a binary classifier $C(x, y)$ which returns 1 if a text $x$ logically
implies a text $y$ and 0 if not, we consider how to define
opinion prevalence of a summary $S$ with sentences $y_1, \ldots, y_n$
with respect to a set of reviews $R = \{ x_1, \ldots, x_m \}$ of
a product $p$.

The opinion prevalence should reward opinions that are expressed by
multiple source reviews.  For a given sentence $y$, this could be formulated
as
\begin{equation}
\frac{1}{\abs{R}}\sum_{i=1}^{\abs{R}} C(x_i, y) \ldotp
\label{eq:support}
\end{equation}
As we consider summaries with $n > 1$ sentences, there are two masks
we want to apply to the classifier values.  One mask should stop us from
counting opinions that were already mentioned; otherwise the one most
frequent opinion could be repeated for a better prevalence than adding
the second most frequent opinion to the summary.  For $y_k$, this mask
looks like $\prod_{j < k} (1 - C(y_j, y_k))$, which becomes zero if
any prior $y_j$ implies $y_k$.

The other mask should stop
us from mentioning conclusions that are so trivial that they follow
from the fact the consumer bought the indicated product, without telling
us anything about the consumer's experience.  Let $t$ be the sentence,
``I bought a $p$.''  For example, if $p$ is a ``Reebok men's basketball
sneaker,'' obvious conclusions like ``It is a shoe'' or ``I wear it'' might
be logically entailed by every single review, even though they are not
interesting.  Therefore we mask $C(x_i, y)$ with $1 - C(t, y)$.

Attaching these masks to equation~\ref{eq:support} and averaging over
sentences in the summary, we obtain our definition of {\em opinion prevalence}:
\begin{eqnarray}
\label{eq:prevalence}
Prev(R, S) & = & \frac{1}{mn} \sum_{k=1}^n \tau_k \rho_k \sum_{i=1}^m C(x_i, y_k)  \\
\tau_k & = & 1 - C(t, y_k) \\
\rho_k & = & \prod_{j < k} (1 - C(y_j, y_k))
\end{eqnarray}
The opinion prevalence is a quantity between 0 and 1.
Based on the results of the previous section, we instantiate the
classifier $C$ with the SummaC MultiNLI document granularity model in the
experiments that follow.
For our ReviewNLI test set, we collected the product names
from the \verb+amazon.com+ and \verb+amazon.ca+ websites using the
item numbers, and used them to compute the trivial masks.
For a dataset where product names are unavailable, it may be necessary
to omit this mask ({\em i.e.} assume $C(t,y) = 0$).

Opinion prevalence is a new quantity, not measured by previously
proposed metrics.  If we assumed that humans could measure prevalence
or that human summaries were optimal, we could compare its correlation
with human judgments to other metrics.  Instead, we have compared the
accuracies with respect to human judgments for various choices of the
classifier $C$, and simply formulated how outputs from the best $C$ 
should be counted.

\section{Opinion Prevalence in Human and Machine Summaries}

Opinion prevalence provides an automatic metric to score the output of
an opinion summarization system, without requiring any reference summaries.
Because of the mental overhead required to remember, compare, generalize, and
count opinions, there is no reason to expect that a human-authored
summary consists of an optimal set of most frequently implied statements.
We investigated the opinion prevalence achieved by the existing human
reference summaries corresponding to the test set of ReviewNLI.
Table~\ref{tbl:summaries} shows the results.

{\bf Do humans extract prevalent opinions any better than extracting
sentences at random?}  We constructed three random summaries for each
product, by concatenating sentences selected at random without replacement
from all the reviews.  To make the length comparable to the human
summaries, we stopped adding sentences after the length in characters
exceeded the length of the first human summary minus the half the length
of its last sentence.  Table~\ref{tbl:summaries} shows a consistent lead
in prevalence for the human summaries over these random summaries.

\begin{table*}[htb]
\begin{center}
\begin{tabular}{lcccccc}
\hline
Method & Original & Original & Original & Simplified & Simplified & Simplified \\
& Prevalence & Characters & Sentences & Prevalence & Characters & Sentences\\
\hline
{\em Abstractive} & & & & & & \\
\hline
Human summary 1 & {\bf .2381} & 288.3 & 4.2 & N/A & N/A & N/A \\
Human summary 2 & .2179 & 297.8 & 4.4 & N/A & N/A & N/A \\
Human summary 3 & .2186 & 302.5 & 4.1 & N/A & N/A & N/A \\
Copycat & .1751 & 157.8 & 3.1 & .2210 & 154.8 & 3.2 \\
COOP & .1999 & 200.6 & 3.3 & .2366 & 197.3 & 4.8 \\
COOP (no search) & .1911 & 196.4 & 3.7 & .2188 & 186.2 & 4.7 \\
RecurSum & .1754 & 195.3 & 4.8 & .1788 & 197.7 & 4.7 \\
\hline
{\em Extractive} & & & & & & \\
\hline
Random summary 1 & .1931 & 299.2 & 5.0 & .1572 & 278.7 & 6.7 \\
Random summary 2 & .1887 & 299.3 & 4.1 & .1682 & 272.1 & 6.8 \\
Random summary 3 & .1791 & 284.4 & 4.6 & .1647 & 275.5 & 7.0 \\
QT & .2039 & 263.5 & 6.1 & .1774 & 248.3 & 9.1 \\
Greedy & {\bf .4744} & 294.8 & 5.3 & {\bf .4886} & 281.5 & 7.4 \\
\hline
\end{tabular}
\caption{Opinion prevalences, lengths in characters, and numbers of sentences
for human summaries, random summaries, and
unsupervised opinion summarization systems.  The simplified columns show
opinion prevalence when simplification preprocessing is added
(see Section~\ref{sec:simp}).
All summary outputs were computed by us
using provided pretrained models.}
\label{tbl:summaries}
\end{center}
\end{table*}

{\bf How do opinion prevalences of existing summarization systems compare?}
Having established these baselines, we measure the opinion prevalence
achieved by various unsupervised opinion summarization systems.
We test three abstractive systems with published models and source code
that were trained on Amazon reviews and evaluated on the Amazon test set
which we used to create ReviewNLI, where aspect information is unavailable:
CopyCat \citep{brazinskas-etal-2020-unsupervised},
COOP \citep{iso-etal-2021-convex-aggregation}, and
RecurSum \citep{isonuma-etal-2021-unsupervised}.
{\bf CopyCat} uses a hierarchical variational autoencoder
\citep{bowman-etal-2016-generating} to encode
reviews, and decodes the mean of their latent vectors, with a decoder that
has access to a pointer-generator mechanism \citep{see-etal-2017-get}.
{\bf COOP} avoids averaging the latent vectors from its autoencoder,
conducting an expensive search over the power set of
source reviews to find a combination of latent vectors that decodes to
a review with best word overlap with the source reviews.
{\bf RecurSum} applies a recursive Gaussian Mixture Model, in which latent
vectors are sampled conditioned on topics, to balance the coverage of summaries
across multiple topics.

We also test an extractive system, {\bf Quantized Transformer (QT)}
\citep{angelidis-etal-2021-extractive}.\footnote{We test QT on Amazon,
but trained per
the repo instructions on 500 entities from SPACE.  Training on Amazon performs
worse than random (.1784) in testing, perhaps suggesting it is hard to learn
a good quantization from the greater variety of products
and fewer reviews per product.}  QT discretizes a transformer
encoding of source review sentences and samples the resulting opinion
clusters proportionately to their popularity.  Therefore, we might expect
it to be well aligned with the objective of opinion prevalence.

We ran each of the four systems on the Amazon test set and measured
the opinion prevalence of the collected outputs, using the provided
precomputed models.  Results are shown in
Table~\ref{tbl:summaries}.

None of the existing systems reaches the range of human performance,
but the gap is not large.  QT performs best, perhaps because popularity
is considered in the sampling procedure.  RecurSum and CopyCat have
worse opinion prevalence than a random summary.  Perhaps the
topic coverage objective of RecurSum is at odds with selecting the
most frequent opinions, regardless of topic.  Among abstractive systems,
only COOP exceeds the prevalence of random summaries.  Without the
search over the power set of input reviews (taking all input reviews
to be selected), its prevalence drops into the range of random summaries.

{\bf Are the results explained by summary length?}
A summarization algorithm that actually optimized for opinion
prevalence may be able to achieve higher prevalences with
summaries of shorter length.  Indeed, reordering sentences and
truncating the least prevalent ones can yield a higher prevalence score:
Suppose $Prev(R, \{y_i\}) \geq Prev(R, \{y_j\}) > 0$ and $C(y_i, y_j) = 0$
for all $i < j$.  Then if $n < n^\prime$, it is easy to show that
$Prev(R, {\{y_k\}}_{k=1}^n) \geq Prev(R, {\{y_k\}}_{k=1}^{n^\prime})$.

However, this phenomenon appears not to explain any of the comparative results
in Table 3.
To the contrary, the existing abstractive systems get lower opinion
prevalence than the human summaries, while having fewer characters and fewer
sentences.  Each instance of simplification
(see Section~\ref{sec:simp}), which preprocesses the input
sentences to be shorter, leads to a greater
number of output sentences, while usually maintaining the same character length
and increasing the opinion prevalence.  Investigating the order of the
implication counts of each statement in the output summaries, we suspect that
only our greedy system would be able to take advantage of this phenomenon to
achieve higher scores, but we have set it to match the character length of the
human and random summaries (which are longest).

\section{Improving Opinion Prevalence}

\begin{table*}[htb]
\begin{center}
\begin{tabular}{p{2cm}p{13cm}}
\hline
Prevalence & Summary \\
\hline
.8000 & I can't speak highly enough about this handy little item! I'd 
recommend this handle highly. It looks good and is built solid.  I
am going to use it with a camera with an active stabilizer. It's
a bit heavy but the construct is very good. \\
.7500 & A real find! It does not irritate my skin and it does the
 job. This does indeed work and if you are not a heavy perspirer it
 will be effective for several days. I can't believe it works so well
and lasts so long. \\
.6667 & Its a great poster! But its perfect for what I needed it
for. Pretty happy with the poster for something to teach my son when
he grows up about just how small our world actually is compared to the
other planets that are out there.  Overall good quality and looks pretty
 good up on the wall.  Big enough to catch someones eye and clear enough
to tell which planet is which. \\
\hline
\end{tabular}
\caption{The three product summaries output by Algorithm 1 with the highest
prevalence.  In the third summary, the last three sentences were regarded
as one by the sentence splitter.}
\label{tbl:greedy}
\end{center}
\end{table*}

\begin{table*}[htb]
\begin{center}
\begin{tabular}{p{2cm}p{13cm}}
\hline
Prevalence & Summary \\
\hline
.7000 & This handle is highly recommended. Would highly recommend it for
lighter cameras. It looks good and is solidly built. A low cost device
reduced jittery videos. It stays pretty secure whether I use it with
the mount or on my flip video camera or snapshot. \\
.6875 & The product is excellent. It does not irritate my skin and it does
the job. It lasts so long. Better product for less money. It works much
better than common over the counter deodorants. It lasts longer than
the deodorant I used to use. \\
.6000 & You have to get the recipe and technique right for tortillas.
This tortilla maker is lovely. A tortilla press needs to be hot to keep
the tortilla from shrinking. Many indicated that users were using this
machine to press and cook tortillas. It takes some trial and errors to
make it work right. \\
\hline
\end{tabular}
\caption{After using an ASSET model as preprocessing, the three product
summaries output by Algorithm 1 with the highest prevalence.}
\label{tbl:simp}
\end{center}
\end{table*}

Because the gap between human, system, and random opinion prevalence is fairly
small, we investigate whether output summaries with much higher prevalence
are possible.

\subsection{Greedy Summaries}

We build a more prevalent extractive summary with a greedy search strategy
over the sentences in the source reviews.  After computing the prevalences
of each statement in the source reviews, we add the most frequently implied
nontrivial statements that are not implied by previously added statements.
The procedure is detailed in Algorithm 1.

\begin{table}[htb]
\begin{center}
\begin{tabular}{l}
\hline
{\bf Algorithm 1}.  Greedy extractive summaries \\
\hline
{\em Input}: Reviews $R = \{x_1, \ldots, x_m\}$ \\
{\em Input}: Trivial statement $t$ \\
{\em Input}: Target minimum length $N$ \\
Collect all sentences $s_1, \ldots, s_n$ from $R$ \\
{\bf foreach} $j = 1, \ldots, n$ \\
\; {\bf if} $C(t_, s_j) = 0$ \\
\; {\bf then} \\
\; \; {\bf foreach} $i = 1, \ldots, m$ \\
\; \; \; Compute $C(x_i, s_j)$ \\
Let $i_1, \ldots, i_n$ be $\{1, \ldots, n\}$ sorted \\
\; so that $Prev(R, \{s_{i_j}\}) \geq Prev(R, \{s_{i_{j+1}}\})$ \\
Let $I \leftarrow \emptyset$ \\
Let $j \leftarrow 0$ \\
{\bf while} $\sum_{i \in I} len(s_i) < N$ and $j < n$ \\
\; Let $j \leftarrow j+1$ \\
\; {\bf if} $C(t, x_{i_j}) = 0$ and $C(x_i, x_{i_j}) = 0$ for $i \in I$ \\
\; {\bf then} Let $I \leftarrow I \cup \{i_j\}$ \\
{\em Output}: Summary $\{s_i\}_{i \in I}$ \\
\hline
\end{tabular}
\label{tbl:algorithm}
\end{center}
\end{table}

We tested this method and found its average opinion prevalence to be
much higher than any system considered, and twice the level of
humans (Table~\ref{tbl:summaries}).

Actual output summaries of Algorithm 1 with the highest prevalence
are shown in Table~\ref{tbl:greedy}.
Qualitatively, they appear nontrivial and nonredundant.  We observed a few
self-contradictory summaries, based on conflicting opinions from
different source reviews, but this problem appeared in the abstractive
and extractive outputs we evaluated too.  (Utilizing the contradiction
classification of the NLI classifier, it may be possible to mitigate
this problem too.)  The complete set of outputs are available in our data
release.

On eight source reviews, this method ran faster than RecurSum
(see Appendix~\ref{sec:computation}), but
as the number of GPU queries grows quadratically with the number of source
reviews, more scalable approaches should be investigated.
We present the method merely to indicate the large potential to
write summaries with better opinion prevalence.

\subsection{Text Simplification}

\label{sec:simp}

Most opinion summarization systems, abstractive or extractive, depend upon
splitting an input review into sentences and considering representations
or properties of each sentence.  In practice, input sentences may be
long and complex, combining many different observations.  This complexity
may make it difficult to extract and relate the common assertions from
different reviews.

To make the job of the summarization system easier, we consider using
a text simplification model to preprocess the input sentences of the
source reviews.  To illustrate the potential of this approach, we
train a T5 base model \citep{t5} on the
ASSET dataset \citep{alva-manchego-etal-2020-asset}.
ASSET collects simplifications from ten human annotators applying a number
of transformations, including rewriting a long sentence as several
shorter sentences.  ASSET source sentences are based on Wikipedia and do not
involve product review data.

ASSET contains only a development and test set, so we use its development set
for training.  The 2,000 development examples provide 20,000 source/target
pairs.  Our model is trained for 30 epochs with maximum source and
target lengths 64, using learning rate $5e^{-5}$, and Adam \citep{adam}
parameters
$\beta_1 = .9$, $\beta_2 = .999$, and $\epsilon = e^{-8}$.

For each sentence in our source reviews, we sample ten sequences
from the ASSET model with nucleus sampling \citep{nucleus} using $p = .9$
and take the first of the outputs with the largest number of sentences.
We apply each of the opinion summarization systems to the source reviews
rewritten by replacing the sentences in this manner.
This preprocessing could be run continually as the source reviews are collected,
before summarization time.

Results are shown in the ``Simplified'' column of Table~\ref{tbl:summaries},
and sample outputs with the highest prevalence are shown in Table~\ref{tbl:simp}.
COOP and CopyCat improve by 3.7\% and 4.6\%, reaching the range of
human performance.  RecurSum changes little, and may be limited more by the
topic tree that structures its output.
We hope these findings encourage the design of opinion summarization
systems based on finer granularities.

Surprisingly, QT and random summaries get worse with preprocessing,
and the greedy summaries from Algorithm 1 improve very little.
After the preprocessing, the outputs of these algorithms can no longer be
considered extractive, and there is no guarantee that a statement will be
supported by at least one review.

\section{Conclusion}

If practitioners are concerned with reflecting a diverse set of the most
common opinions in an opinion summary, we recommend that they evaluate
opinion prevalence.  Mimicking human summaries will not draw systems
nearer to this goal because humans achieve less than half the prevalence
that is possible.  Some systems can make easy progress by replacing
their use of sentences with finer grained information;
preprocessing with a simplification model achieves this with no changes
to the underlying system.

Sometimes opinion summaries are expected to cover particular aspects,
such as atmosphere, service, and food at a restaurant, or to balance
positive and negative opinions.
Although we do not consider this setting here, our metric could
be used restricted to statements relevant to each aspect or sentiment.

No single metric captures all desiderata of an opinion summary.
We recommend opinion prevalence be used
in combination with other metrics to evaluate attributes
such as fluency, coherence, and informativeness.

We introduced a greedy summarizer (Algorithm 1) to show the existence of
summaries with twice the opinion prevalence of humans, though a linear time
construction of them remains a challenge for future research.
The success of the text simplification preprocessing gives an easy improvement
which could be added to systems designed to embed opinions at the
sentence granularity.

\section*{Limitations}

Although triviality masking provides some protection against generic
statements, not every uninformative statement will be
implied by the trivial statement.  Rankings of systems by opinion prevalence
still should be supplemented by human assessments of informativeness
and redundancy.

We have only evaluated opinion prevalence on one dataset, but we expect
opinion prevalence of human summaries to be even worse on datasets with more
source reviews per summary.

\section*{Ethics Statement}

An opinion summarization system poses the risk that a consumer may
select a product while missing information that they could have found if
they read more of the reviews directly.  Such a system also may hallucinate
or pose misleading summaries of content.  The summary may not represent
the most prevalent opinions, though the present work provides a means
to ameliorate that risk.  

On the positive side, opinion summaries might provide information from
more reviews than the consumer could have read in a short amount of time.
It is in this hope that we are developing technologies
for such summarization systems.

Consumers may want to search reviews for tail risks, such as a product
catching on fire.
Because (hopefully) these events are uncommon experiences,
a summarization method targeting prevalent opinions may overlook such reports.
Faults of a product might be found in a body of mostly positive reviews
by conditioning on negative sentiment before applying the method.
For particular risks, simply using an NLI classifier to compare
product reviews to the statement of that risk might expose those
risk possibilities.


\bibliographystyle{acl_natbib}
\bibliography{anthology,kdd2023}


\appendix

\section{Crowdworker Protocol}

\label{sec:crowdworker}

We recruited crowdworkers from
Amazon Mechanical Turk.\footnote{{\tt www.mturk.com}}
Workers were required to live in the US, Great Britain, or Australia,
have a 90\% approval rate, and have 1,000 approved HITs.
We further restricted the workers to have demonstrated satisfactory
performance on our previous HITs.  Additionally, they had to classify
seven out of eight examples on a qualification test
in agreement with our answers.

Workers were compensated 40 cents for each HIT, which consisted of
a single source review and a single opinion summary.
For each sentence in the summary, the workers judged whether
the sentence was ``mostly or fully supported by the review''
or ``partially or totally unsupported by the review'' (a binary decision).
A rough estimate of the average time a worker spent on each assignment
is 90 seconds, so that crowdworkers could earn \$16 per hour, which is above
minimum wage in every state of the United States
except the District of Columbia.\footnote{{\tt www.statista.com/statistics/238997/\\minimum-wage-by-us-state}}

Detailed instructions were as follows:

\begin{quotation}
You're given an Amazon review.  Decide whether the summary statements
given afterwards are mostly supported by the review you read, or not.
If significant parts of the summary statement are not supported
({\em e.g.} the summary says ``The hotel room was clean and bright'' and
the review just says ``We liked the clean rooms''), select ``partially
unsupported.'' Please see the examples for guidance.

{\em Example 1}: These are the perfect tights for my 5-year old. The tights are very well made and have already lasted several washings (hang dry). The color is beautiful, and my daughter loves that she can wear flip-flops to class like the big girls do.

{\em Summary statements}:
\begin{itemize}
\item These tights are great. - SUPPORTED
\item They are durable and do not tear easily, they can be worn and washed
without worry. - SUPPORTED - {\em Washing and durability are discussed,
and not tearing is part of durability.}
\item The bottoms of can be pulled up easily so that sandals can be worn
with them. - SUPPORTED - {\em The review explained that flip flops can
be worn with them, probably for the same reason.}
\item It might be a good idea to order a size bigger because they can be
a little tight in the waist. - NOT SUPPORTED
\item Overall, these tights are definitely recommended - SUPPORTED -
{\em Inferred from the reviewer's strong satisfaction.}
\end{itemize}

{\em Example 2}: My son is 3 and this fits him perfectly.
He'll probably be able to wear it for the next two years if he'd like.
It's cute too.  The hat is thin, but completes the outfit.  And the
candy pocket is huge.  Perfect!  I'm so glad we bought this costume over
any other Thomas costume.

{\em Summary statements}:
\begin{itemize}
\item What an impressive Thomas the Train costume! - SUPPORTED - {\em The
reviewer certainly seems impressed.  We aren't quite sure about whether
it's a ``Thomas the Train'' costume, but we allow that it probably is on the
basis of ``Thomas costume.''}
\item It will fit any train loving toddler for at least a few Halloween
seasons due to its roomy design. - SUPPORTED - {\em The reviewer mentions
expecting it to work for at least two more years, so it must have
growing room.}
\item The candy pocket on the front is an adorable touch. - NOT SUPPORTED
- {\em The reviewer mentions the candy pocket but says it is huge,
which is different from adorable}
\item And it's so stinkin' cute! - SUPPORTED - {\em ``It's cute too.''}
\item Absolutely worth the price. - NOT SUPPORTED - {\em The review
doesn't mention price or value.  It may have been very expensive.}
\end{itemize}
\end{quotation}

\section{Artifacts}

The development and test sets of Amazon reviews and human summaries used
for evaluation are distributed with Copycat by \citet{brazinskas-etal-2020-unsupervised}
under the MIT license.\footnote{{\tt github.com/abrazinskas/Copycat-\\abstractive-opinion-summarizer}}
The upstream Amazon data is not clearly licensed, but we believe our research
use is consistent with previous research uses of the data,
including \citet{brazinskas-etal-2020-unsupervised},
\citet{iso-etal-2021-convex-aggregation}, and
\citet{isonuma-etal-2021-unsupervised}.
The ReviewNLI logical consistency judgments will also be
distributed under the MIT license.

Among consistency metrics, SuperPAL\footnote{{\tt github.com/oriern/SuperPAL}}
\citep{ernst-etal-2021-summary}
and SummaC\footnote{{\tt github.com/tingofurro/summac}}
\citep{laban-etal-2022-summac}
are Apache 2.0 licensed.
QAFE\footnote{{\tt github.com/salesforce/QAFactEval}}
\citep{fabbri-etal-2022-qafacteval}
is BSD licensed.
ROUGE is run by the Huggingface Datasets library
\footnote{{\tt github.com/huggingface/datasets}}
which wraps the Google Research implementation\footnote{{\tt github.com/google-research/\\google-research/tree/master/rouge}}
both of which are Apache 2.0 licensed.

Among summarization systems, Copycat,
RecurSum\footnote{{\tt github.com/misonuma/recursum}}
\citep{isonuma-etal-2021-unsupervised}, 
and QT\footnote{{\tt github.com/stangelid/qt}}
\citep{angelidis-etal-2021-extractive}
are MIT licensed.  COOP\footnote{{\tt github.com/megagonlabs/coop}}
\citep{iso-etal-2021-convex-aggregation}
is BSD licensed.

The worker ID's are unique identifiers of the Amazon crowdworkers
who contributed the opinion consistency judgments, and we will
apply a one-way hash to them before releasing them in ReviewNLI.
The workers did not provide any free text responses.

\section{Computational Resources}

\label{sec:computation}

Experiments were run on Nvidia GeForce GTX 1080Ti GPU's.
Unless otherwise noted, models were pretrained and one GPU was
used for evaluation.

For the consistency metrics, QAFE uses a BART large \citep{lewis-etal-2020-bart}
model for question generation (400M parameters) and an Electra large
\citep{electra} model for question answering (335M parameters).
SuperPAL uses a Roberta large model (355M parameters) for proposition
alignment, and we suspect its OpenIE model for proposition extraction
has about 1.8M parameters based on file size.
The SummaC model that performed best was Roberta large (355M parameters).

For the summarization systems, we counted a total of 41M parameters
in the Copycat checkpoint, 21M parameters in the COOP checkpoint,
and 23M parameters in the RecurSum checkpoint.
Using one GTX 1080 Ti GPU, we trained the 27M parameter QT model on the
1.1M reviews (for 11K hotels) in the SPACE corpus
\citep{angelidis-etal-2021-extractive}
as instructed, in just under four hours.

Running the consistency metrics between each input review and the first human
summary on the 32 product test set using the same GPU, wall clock
run times (including latency) were as follows:
SuperPAL took 6 minutes and 9 seconds; QAFE took 3 minutes and 3 seconds;
and SummaC took 1 minute and 23 seconds.  The opinion prevalence
scores of the first human summaries of the 32 products were
computed in 33 seconds, which is faster
than SummaC because of early stopping when a trivial or
previously implied statement is encountered.

Running the summarization systems on the test set, wall clock run times
were as follows: Copycat took 24 seconds; COOP took 1 minute 36 seconds;
RecurSum took 5 minutes and 17 seconds; and QT took 10 seconds.
Computing the greedy extractive summaries (Algorithm 1) took
2 minutes 48 seconds.

For the ASSET simplification model (297M parameters),
30 epoch training was chosen due to the small training (development) set
after artifacts (insertions of nonsense words) were observed in the output of
a 3 epoch model, but other training parameters were left at Huggingface
defaults and no careful model selection was performed.

\end{document}